\documentclass[conference]{IEEEtran}
\IEEEoverridecommandlockouts
\usepackage{cite}
\usepackage{amsmath,amssymb,amsfonts}
\usepackage{algorithmic}
\usepackage{graphicx}
\usepackage{textcomp}
\usepackage{xcolor}
\def\BibTeX{{\rm B\kern-.05em{\sc i\kern-.025em b}\kern-.08em
    T\kern-.1667em\lower.7ex\hbox{E}\kern-.125emX}}

\usepackage{algorithm}
\usepackage{algorithmic}
\usepackage{threeparttable}    
\usepackage{multirow}

\usepackage{orcidlink}
\usepackage{lipsum}
\usepackage{fancyhdr}
\fancypagestyle{sec}{\lhead{\tmpx}}

\fancypagestyle{arXiv}
{
   \fancyhf{}
   \chead{\textcolor{gray}{This paper has been accepted for publication at the 2023 IEEE Biomedical Circuits and Systems (BioCAS) Conference, Toronto, Canada}}
   \cfoot{ \fontsize{=9pt}{10pt}\selectfont  \textcolor{gray}{© 2023 IEEE. Personal use of this material is permitted. Permission from IEEE must be obtained for all other uses, in any current or future media, including reprinting/republishing this material for advertising or promotional purposes, creating new collective works, for resale or redistribution to servers or lists, or reuse of any copyrighted component of this work in other works}\\\fontsize{=10pt}{12pt}\selectfont\thepage}
   
}

\begin{document}

\title{3ET: Efficient Event-based Eye Tracking using a Change-Based ConvLSTM Network

}

\author{
\IEEEauthorblockN{
Qinyu Chen\IEEEauthorrefmark{1} \orcidlink{0009-0005-9480-6164}, \IEEEmembership{Member~IEEE}, Zuowen Wang\IEEEauthorrefmark{1} \orcidlink{0000-0002-8051-9886}, Shih-Chii Liu\IEEEauthorrefmark{1}, \IEEEmembership{Fellow~IEEE}, Chang Gao\IEEEauthorrefmark{2} \orcidlink{0000-0002-3284-4078}, \IEEEmembership{Member~IEEE}
}
\IEEEauthorblockA{\IEEEauthorrefmark{1}Institute of Neuroinformatics, University of Zurich and ETH Zurich, Switzerland}
\IEEEauthorblockA{\IEEEauthorrefmark{2}Department of Microelectronics, Delft University of Technology, Netherlands}
\vspace{-1cm}
\thanks{
Corresponding Author: Chang Gao}
\thanks{
 This work was funded by the Swiss National Science Foundation BRIDGE - Proof of Concept Project (40B1-0\_213731). }
}



\maketitle

\begin{abstract}

This paper presents a sparse Change-Based Convolutional Long Short-Term Memory (CB-ConvLSTM) model for event-based eye tracking, key for next-generation wearable healthcare technology such as AR/VR headsets. We leverage the benefits of retina-inspired event cameras, namely their low-latency response and sparse output event stream, over traditional frame-based cameras. Our CB-ConvLSTM architecture efficiently extracts spatio-temporal features for pupil tracking from the event stream, outperforming conventional CNN structures. Utilizing a delta-encoded recurrent path enhancing activation sparsity, CB-ConvLSTM reduces arithmetic operations by approximately 4.7$\times$ without losing accuracy when tested on a \texttt{v2e}-generated event dataset of labeled pupils. This increase in efficiency makes it ideal for real-time eye tracking in resource-constrained devices. The project code and dataset are openly available at \url{https://github.com/qinche106/cb-convlstm-eyetracking}.

\end{abstract}

\begin{IEEEkeywords}
Pupil tracking, event cameras, sparsity, ConvLSTM, healthcare, AR/VR.
\end{IEEEkeywords}

\section{Introduction}
\thispagestyle{arXiv}
\IEEEPARstart{T}{he} process of eye movements often reveals our mental processes and comprehension of the visual realm. Implementing eye tracking technology offers many possibilities in augmented reality/virtual reality (AR/VR) domains, enabling techniques like foveated rendering to offer a more compelling and efficient visual experience~\cite{fernandes2023leveling,fuhl2021teyed}. 
Eye tracking has potential benefits in wearable healthcare applications. For instance, it can aid in identifying eye movement disorders associated with diseases like Parkinson's or Alzheimer's, thereby enabling early diagnosis and regular assessments~\cite{pretegiani2017eye,duan2019dataset}. 

Considering the energy and computational constraints of mobile headsets and the high sampling rate needed for applications like predictive foveated rendering for enhanced VR experiences, the eye tracking system should be lightweight to facilitate low-power and low-latency operation~\cite{Eventgaze2021}. Eye tracking systems relying on high-speed and high-resolution cameras are costly and power-intensive. Notably, most of the image is redundant in near-eye eye tracking as the only significant movement typically originates from the eye region, with the rest remaining static.

\begin{figure}[t]
    \centering
    \includegraphics[width=0.35\textwidth]{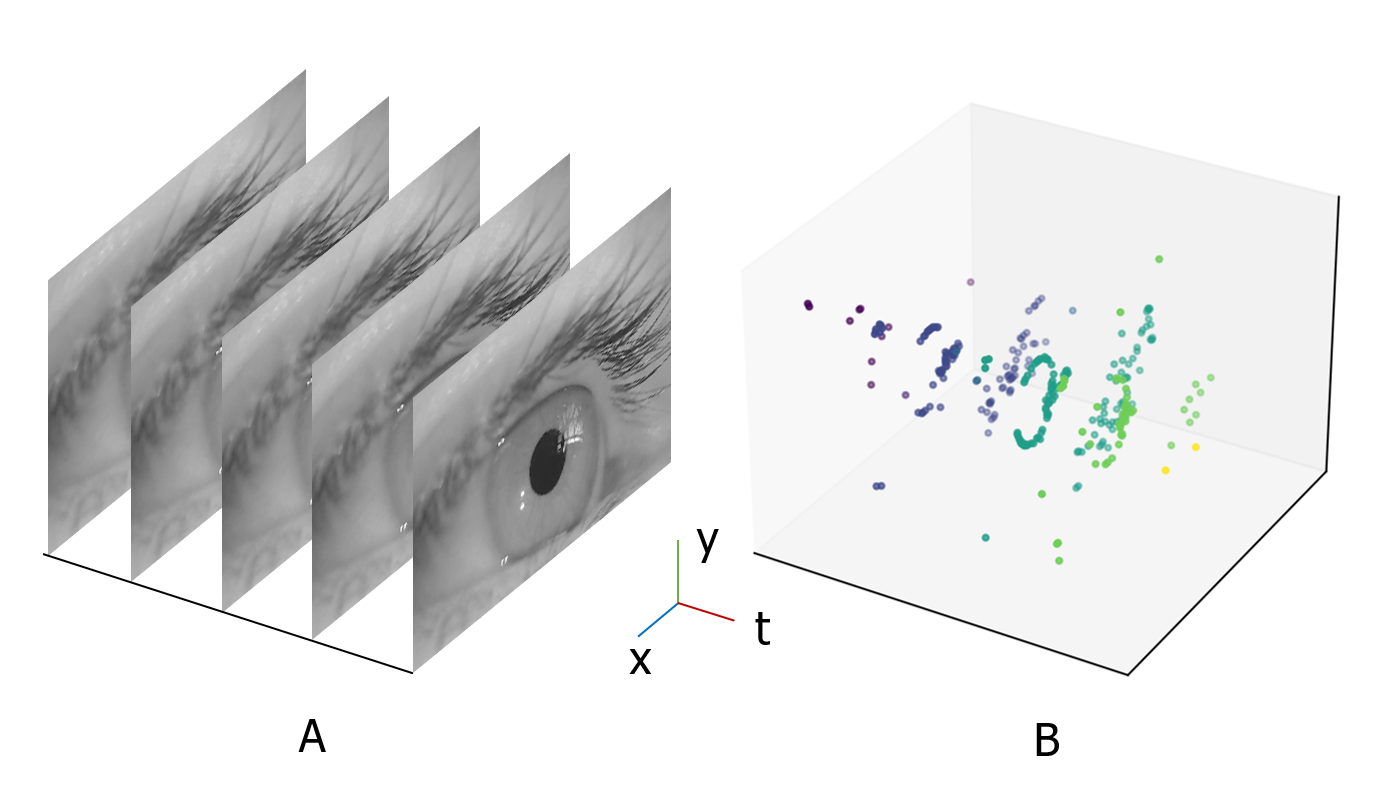}
    \caption{Comparison between frames and events for the same 53~ms eye movement motion. A) Example video from the LPW dataset \cite{tonsen16_etra}. Frames are sampled at a fixed frame rate (95~kHz); B) 
    Using the \texttt{v2e} simulator \cite{hu2021v2e}, the video frames in A are converted to realistic synthetic DVS event streams. In this example, 5 frames of size 240$\times$180 produce only 310 events.}
    \label{fig:comp}
    \vspace{-4mm}
\end{figure}

Dynamic Vision Sensors (DVS) or event-based cameras, which capture brightness changes as sparse events, emerge as a viable solution for near-eye tracking (Fig.~\ref{fig:comp}). The sparsity in the DVS events originating from eye movements or pupil size changes can significantly reduce computational demands compared to conventional camera-based systems. Furthermore, as event-based cameras only record changes in brightness levels, they protect the user's privacy by avoiding the collection of detailed iris data.

Eye tracking is a significant field in computer vision~\cite{lee2020deep, eivazi2019improving, liu20203d}, yet it's relatively unexplored with event cameras due to the scarcity of relevant event-based datasets~\cite{ryan2021real, feng2022gaze}. Two common approaches guide recent advances in event-based eye tracking algorithms, mirroring those of traditional computer vision: (1) The 3D model-based method locates key points corresponding to the image's geometrical features and fits them to a 3D eye model using optimization techniques. Despite their accuracy, these methods have limitations in resource-limited platforms like headsets, which often require frequent user-specific calibrations. (2) The appearance-based method employs Convolutional Neural Networks (CNNs) to track the eye within the raw image. However, CNNs tend to isolate spatial features, treating input data as independent and often ignoring the potential temporal context in the data sequence.
Our study is aligned with the second approach, emphasizing pupil center detection, an integral component of eye tracking techniques.

This paper introduces an Efficient Event-based Eye Tracking (3ET) solution that integrates a ConvLSTM architecture merging convolution operators with recurrent units, thereby enhancing its efficiency in extracting sparse spatio-temporal features from event streams. We also propose a Change-Based ConvLSTM (CB-ConvLSTM) network to alleviate the computational burden. 
This innovative network introduces high sparsity into the process without compromising performance.
Experiments demonstrate that our approach yields more than 30\% higher accuracy than the CNN-based model while achieving a 4.7$\times$ reduction in arithmetic operations without any accuracy loss compared to the standard ConvLSTM-based model when used on an event-based pupil dataset.

\begin{figure}[t]
    \centering
    \includegraphics[width=0.38\textwidth]{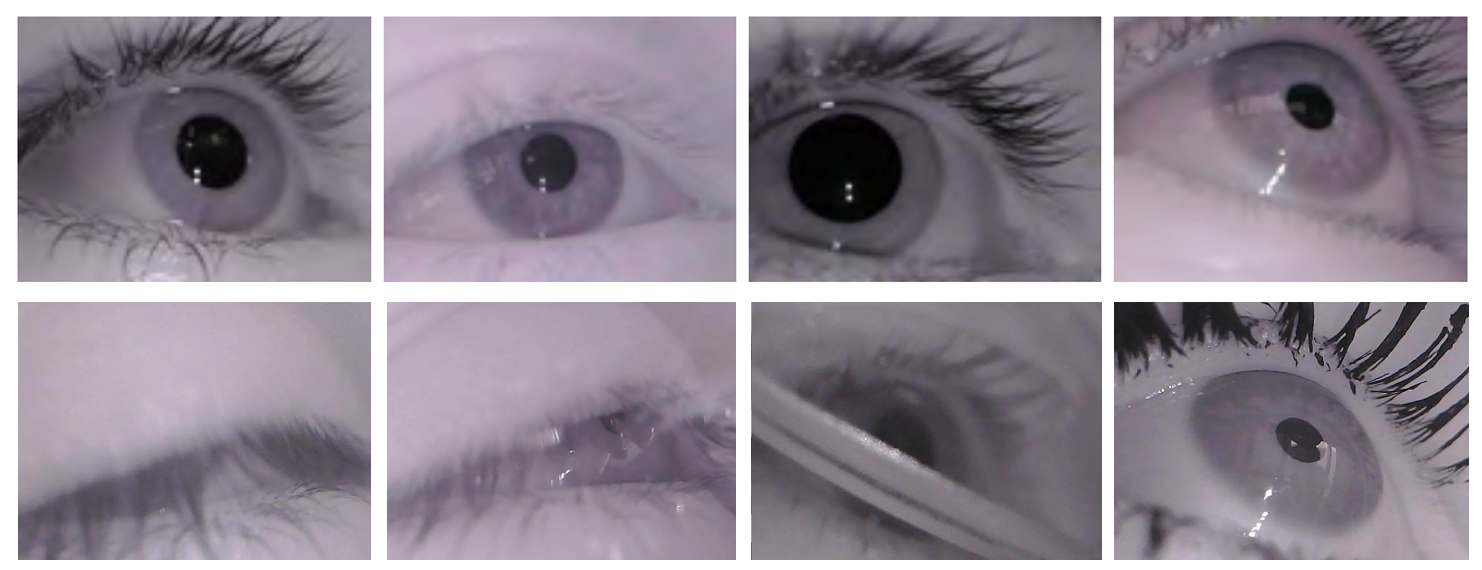}
    \caption{Diverse set of images from LPW dataset~\cite{tonsen16_etra}. The first row shows different eye appearances. The second row shows some difficult cases, e.g. eyelid occlusion, glasses occlusion, and heavy makeup. 
}
    \label{fig:eye}
    \vspace{-2mm}
\end{figure}

\begin{figure}[t]
    \centering
    \includegraphics[width=0.38\textwidth]{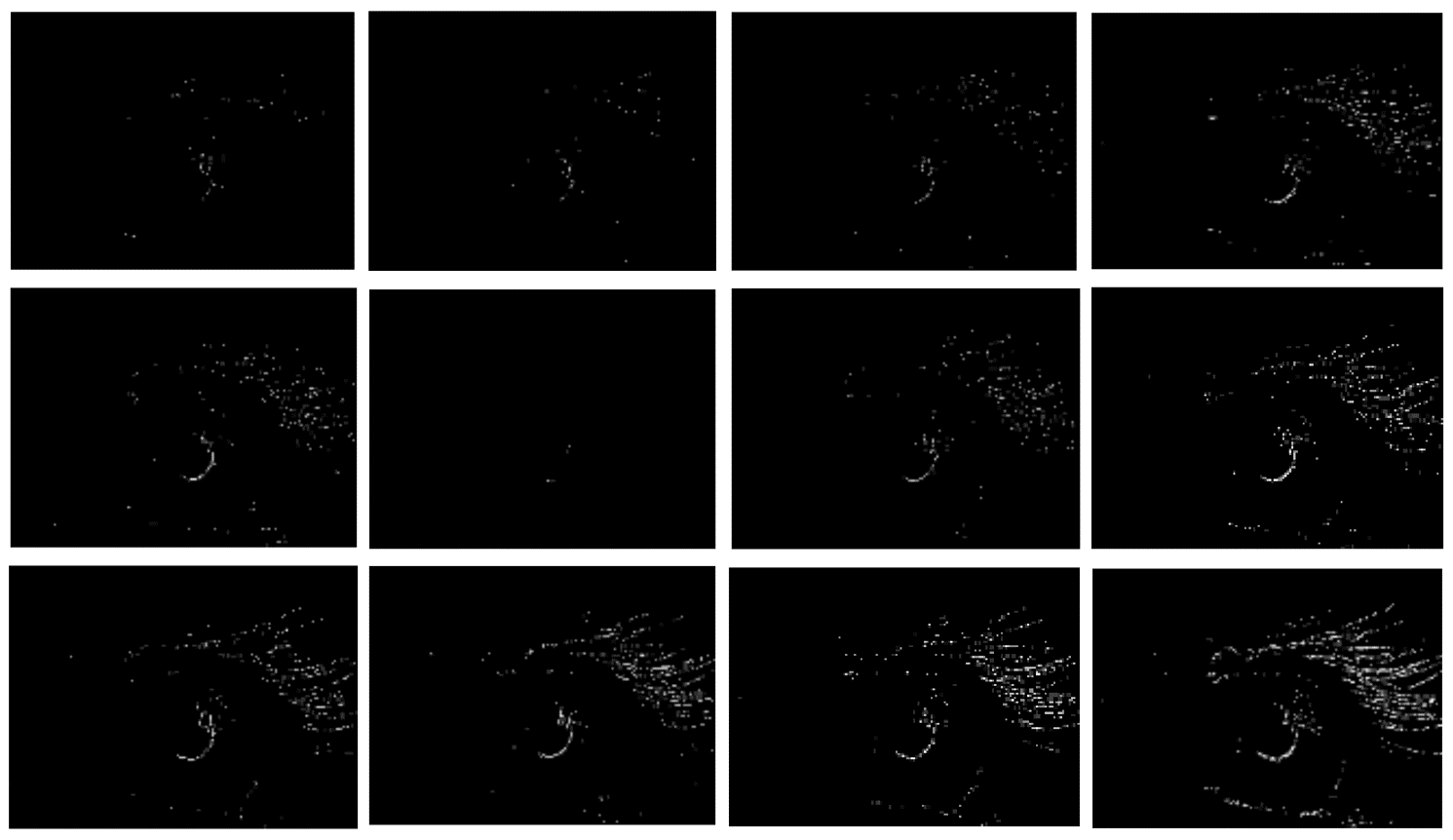}
    \caption{A set of continuous event-based frames using voxel grid representation from event-based LPW dataset using DVS simulator \texttt{v2e} tool.}
    \label{fig:eventeye}
    \vspace{-2mm}
\end{figure}

\section{Methods}


A DVS pixel triggers an event in response to a localized change in brightness, yielding a sparse event stream in certain scenes. This inherent sparsity of input data presents opportunities for improved processing speed and heightened computational efficiency within postprocessing algorithms. 
The primary aim of our work is to devise a cost-efficient algorithm for pupil detection and tracking that is friendly for real-time inference by effectively exploiting the intrinsic sparsity in event data streams. We first outline the ConvLSTM architecture that is implemented to process this temporal data stream. Subsequently, we describe our proposed CB-ConvLSTM architecture, designed specifically to reduce the computational overhead of the network.

\subsection{Dataset: Event-based Dataset for Eye Movement Tracking}

Owing to the non-availability of a DVS dataset tailored specifically for eye tracking applications, we resort to \texttt{v2e}~\cite{hu2021v2e}, an open-source DVS simulator, to transform an existing RGB dataset, namely Labeled Pupils in the Wild (LPW) \cite{tonsen16_etra}, into a synthetic dataset composed of DVS event streams. The LPW dataset includes 66 high-quality videos capturing the eye region, each spanning approximately 20 seconds. Several representative examples are displayed in Fig.~\ref{fig:eye}. To circumvent instances where no events are produced over a prolonged period, we selectively incorporated one-third of the video clips that show above-average speed in eye movements.

The produced event streams, depicted as a sequence of events, are stored as \texttt{.h5} files within the \texttt{v2e} simulator. Each $\textit{event}$, marked by index $i$ in an event stream, is expressed as $e_i =(x_i, y_i, t_i, p_i)$, where $(x_i, y_i)$ signifies the pixel location, $t_i$ is the timestamp, and $(p_i = \pm 1)$ represents the polarity or the direction of the brightness change. In this study, we employ the constant time-bin count representation~\cite{survey}, indicated as $V(x,y)$, where $(x, y)$ is the pixel location.

The events occurring within a time window $\Delta T$ are translated into a $H \times W$ frame, with $H$ and $W$ representing the height and width of the event-based frame, respectively. This transformation is illustrated in the equation below,
\begin{equation}
    V(x,y)= \sum p_i* I(x, y, t_i, x_i, y_i, T1, T1 + \Delta T) 
\end{equation}
where the summation includes all events $e_i$ in the event stream, and $I(x, y, t_i, x_i, y_i, T1, T1 + \Delta T)$ is the indicator function that equals to 1 when $x=x_i$, $y=y_i$ and $T1 < t_i \leq T1 + \Delta T$. The function $V(x, y)$ yields the accumulated value of all events that occur within the time window from $T1$ to $T1 + \Delta T$ at each location $(x, y)$.

For the purpose of this research, we establish $\Delta T$ at 4.4 ms to ensure synchronicity with the frame rate of the source RGB dataset. This synchronization guarantees the precise alignment of the event frame with the corresponding labels. Within the simulator, the initial 640$\times$480 frames are converted to the 240$\times$180 size output of the DAVIS240~\cite{dvs2008}, and the resultant DVS output is further resized to 80$\times$60 to reduce network training time. The final synthetic event-based dataset comprises 11k event-based frames derived from the 22 videos. Fig.~\ref{fig:eventeye} exhibits a selection of continuous event-based frames from a video sequence.

\begin{figure*}[t]
    \centering
    \includegraphics[width=0.95\textwidth]{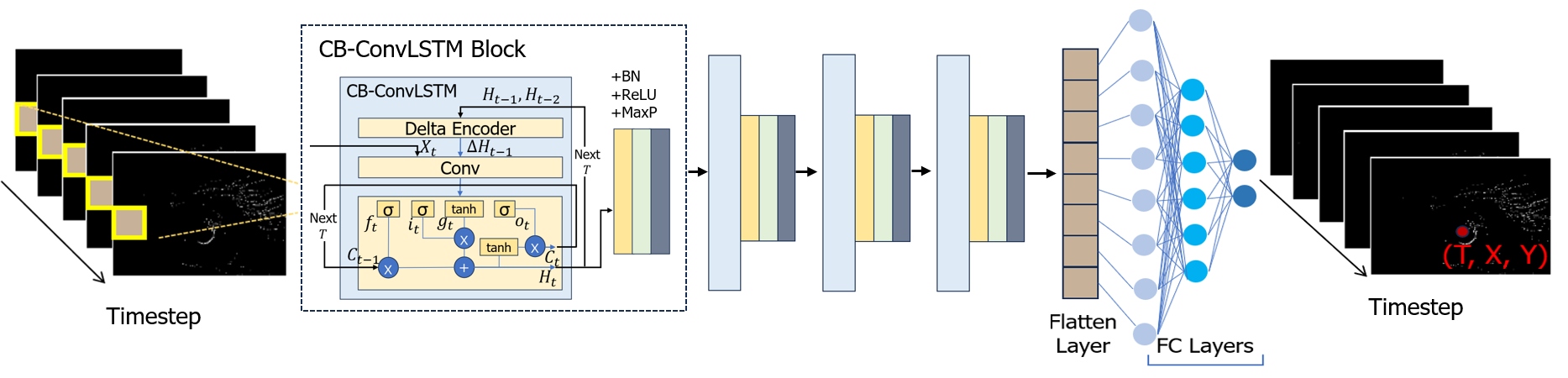}
    \caption{The pupil tracking network using Change-Based ConvLSTM (CB-ConvLSTM) units on event-based LPW dataset}
    \label{fig:convlstm}
    \vspace{-5mm}
\end{figure*}




\subsection{Pupil tracking using ConvLSTM Network on event-based LPW dataset}
\label{Pupil}
In previous work, deep CNN models \cite{lee2020deep,eivazi2019improving} were used to detect the pupil center. The networks performed reasonably well on the RGB dataset. As can be observed from Fig.~\ref{fig:eye}, the pupil in the image has a distinct outline in most cases, which aids pupil detection.
However, the model faces challenges when processing the sparse event-based frames depicted in Fig.~\ref{fig:eventeye}. Many event-based frames lack sufficient events to enable reliable 
prediction on a per-frame basis.
We suggest integrating recurrent structures like LSTM units into the neural network to tackle this issue. Such structures are better suited for interpreting temporal information across the sequence of eye movements in a video, thus improving the prediction accuracy on sparse, event-based frames.

However, the conventional LSTM is computationally expensive because of the all-to-all connections in the architecture.  
It does not inherently preserve or encode any spatial information. 
To solve this problem, the ConvLSTM \cite{shi2015convolutional} was proposed to additionally capture the spatial dependencies in the data.
In this work, the ConvLSTM is used to capture spatial-temporal dependency in the sparse event-based frame stream. 

A ConvLSTM unit is composed of an input gate $i$, a forget gate $f$, a cell gate $g$, an output gate $o$, and a memory cell state $c$. 
Gates $i$, $f$, and $g$ control the update of the cell $c$ state.
Gate $o$ determines the proportion of cell memory that is transferred to the hidden state output $h$. 
The update equations of a ConvLSTM layer are given as:
\begin{equation}
    \begin{aligned}
    	\mathbf{i}_{t} &=\sigma\left(\mathbf{W}_{xi}*\mathbf{X}_t + \mathbf{W}_{hi}*\mathbf{H}_{t-1} + \mathbf{b}_{i}\right) \\
    	\mathbf{f}_{t} &=\sigma\left(\mathbf{W}_{xf}*\mathbf{X}_t + \mathbf{W}_{hf}*\mathbf{H}_{t-1} + \mathbf{b}_{f}\right) \\
    	\mathbf{g}_{t} &= \tanh\left(\mathbf{W}_{xg}*\mathbf{X}_t + \mathbf{W}_{hg}* \mathbf{H}_{t-1} + \mathbf{b}_{g}\right) \\
     \mathbf{o}_{t} &=\sigma\left(\mathbf{W}_{xo}*\mathbf{X}_t + \mathbf{W}_{ho}*\mathbf{H}_{t-1} + \mathbf{b}_{o}\right) \\
            \mathbf{C}_{t} &=\mathbf{f}_{t}\odot \mathbf{C}_{t-1}+\mathbf{i}_{t}\odot \mathbf{g}_{t} \\
            \mathbf{H}_{t} &=\mathbf{o}_{t}\odot \tanh{(\mathbf{C}_{t})}
    \end{aligned}
    \label{eq:lstm}
\end{equation}
where $*$ and $\odot$ signify the convolution and Hadamard functions, respectively. $\mathbf{X}_t$ is the input video frame tensor, $\mathbf{H}_t$ is the hidden state tensor, $\mathbf{C}_t$ is the memory cell tensor, $\mathbf{W}$ denotes weight matrices, and $\mathbf{b}$ denotes bias terms.
 
The model used for the pupil tracking dataset consists of four ConvLSTM layers and two FC layers, with a total of $\sim$0.42 million parameters.
The ConvLSTM layers comprise 8, 16, 32, and 64 hidden nodes, respectively, using a kernel size of 3$\times$3. The outputs of each ConvLSTM layer go through a batch normalization function and a ReLU activation, followed by a max pooling layer for downsampling.
The first FC layer has 128 hidden neurons, and the second FC layer generates two outputs corresponding to the pupil center's x and y coordinates.
Note that the operations in the FC layers will be executed $T$ times, where $T$ signifies the length of the input sequence in the time dimension.

\subsection{Change-Based ConvLSTM (CB-ConvLSTM) Network for Inducing Activation Sparsity}



The inherent sparsity of the input event streams offers a significant opportunity to curtail computational costs related to network processing, as was studied in previous works~\cite{wang22transformer, Messikommer20eccv}. An in-depth evaluation of each network stage underscores the fact that within a ConvLSTM block, convolutions are the most computation-demanding modules.

Within every ConvLSTM block, the input consists of two elements: the input tensor $\mathbf{X}_t$ and the hidden state $\mathbf{H}_{t-1}$. The first component, $\mathbf{X}_t$, is naturally sparse, attributed to its origin from event-based frames in the context of the first ConvLSTM layer and due to the application of the ReLU activation in subsequent ConvLSTM layers. In contrast, the second input part, $\mathbf{H}_{t-1}$, is predominantly dense.

In light of this, we introduce Change-Based ConvLSTM (CB-ConvLSTM), aimed at inducing sparsity into hidden states. Instead of employing $\mathbf{H}_{t-1}$, we utilize the change between $\mathbf{H}_{t-1}$ and $\mathbf{H}_{t-2}$ as the recurrent input feature. Additionally, we set a threshold $\theta$ for this change to foster increased sparsity. The formulation of the thresholded change $\Delta \mathbf{H}_{t-1}$ is presented below:

\begin{equation}
\label{eq6}
\Delta \mathbf{H}_{t-1}=\left\{
\begin{aligned}
\left(\mathbf{H}_{t-1} - \mathbf{H}_{t-2} \right) & , & \left(\mathbf{H}_{t-1} - \mathbf{H}_{t-2} \right) \geq \theta, \\
0 & , & \left(\mathbf{H}_{t-1} - \mathbf{H}_{t-2} \right) < \theta.
\end{aligned}
\right.
\end{equation}
The equations of a CB-ConvLSTM unit are shown below:
\begin{equation}
    \begin{aligned}
    	\mathbf{i}_{t} &=\sigma\left(\mathbf{W}_{xi}*\mathbf{X_t} + \mathbf{W}_{hi}*\mathbf{\Delta H}_{t-1} + \mathbf{b}_{i}\right) \\
    	\mathbf{f}_{t} &=\sigma\left(\mathbf{W}_{xf}*\mathbf{X}_t + \mathbf{W}_{hf}*\mathbf{\Delta H}_{t-1} + \mathbf{b}_{f}\right) \\
    	\mathbf{g}_{t} &= \tanh\left(\mathbf{W}_{xg}*\mathbf{X}_t + \mathbf{W}_{hg}* \mathbf{\Delta H}_{t-1} + \mathbf{b}_{g}\right) \\
     \mathbf{o}_{t} &=\sigma\left(\mathbf{W}_{xo}*\mathbf{X}_t + \mathbf{W}_{ho}*\mathbf{\Delta H}_{t-1} + \mathbf{b}_{o}\right) \\
            \mathbf{C}_{t} &=\mathbf{f}_{t}\odot \mathbf{C}_{t-1}+\mathbf{i}_{t}\odot \mathbf{g}_{t} \\
            \mathbf{H}_{t} &=\mathbf{o}_{t}\odot \tanh{(\mathbf{C}_{t})}
    \end{aligned}
    \label{eq:cbconvlstm}
\end{equation}

This modification induces a high level of temporal sparsity in convolutions considering that $\Delta \mathbf{H}_{t-1}$ comprises zero values. This approach sets itself apart from preceding change-based networks~\cite{neil2017deltanet,gao2022spartus} by exclusively employing a delta encoder for the hidden path, thereby circumventing the necessity to accumulate previous matrix-vector multiplication results. As a result, both computational and memory overhead associated with inducing and leveraging temporal sparsity is significantly reduced.
Leveraging the CB-ConvLSTM unit as a base, we construct a network for pupil tracking, maintaining the same topology with the network described in Section~\ref{Pupil} as illustrated in Fig.~\ref{fig:convlstm}.

\begin{table}[t]\renewcommand\arraystretch{1.0}
\begin{threeparttable}[b]
\caption{Sparsity level in CB-ConvLSTM and vanilla ConvLSTM}
\label{sparsity}
\begin{center}
\setlength{\tabcolsep}{1.6mm}{
    \begin{tabular}{c|ccccc|c}
\hline
\hline
&\multicolumn{5}{c|}{CB-ConvLSTM} & ConvLSTM \\
\multicolumn{1}{c|}{$\theta$} && 0.5   & 0.2 & 0.1 & 0  & - \\\hline
\multirow{3}{*}{ConvLSTM 1}&Inp. sp\footnotemark[1]& 0.955 & 0.955  & 0.955 &   0.955   &0.955 \\
                        &Hid. sp& 0.999 & 0.999  & 0.998 &   0.733   &0.244 \\ \hline
\multirow{3}{*}{ConvLSTM 2}&Inp. sp& 0.606 & 0.630  & 0.636 &   0.767   &0.672 \\
                        &Hid. sp& 0.998 & 0.994  & 0.969 &   0.804  &0.179  \\\hline
\multirow{3}{*}{ConvLSTM 3}&Inp. sp& 0.596 & 0.589  & 0.541 &   0.735   &0.572 \\
                        &Hid. sp& 0.997 & 0.969  & 0.928 &   0.677   &0.195  \\\hline
\multirow{3}{*}{ConvLSTM 4}&Inp. sp& 0.543 & 0.424  & 0.465 &   0.604   &0.500 \\
                        &Hid. sp& 0.989 & 0.934  & 0.878 &   0.623   &0.246 \\\hline
\multirow{3}{*}{       }&Inp. sp& 0.763 & 0.760  & 0.754 &   0.850   &0.778 \\
                        &Hid. sp& 0.998 & 0.988  & 0.969 &   0.738   &0.216  \\
                        &Tot. sp& 0.924 & 0.916  & 0.903 &   0.750   &0.330\\\hline\hline
\multirow{1}{*}{Network}&\textbf{Tot. sp}& 0.853 & 0.845  &  0.833 &   0.692  &0.304\\ 
                        \hline 
\end{tabular}}
\end{center}
\begin{tablenotes}
       \footnotesize
       \item[1]Inp. sp, Hid. sp, Tot. sp denote the input tensor sparsity, the hidden state tensor sparsity, and the total sparsity. 
\end{tablenotes}
\end{threeparttable}
\end{table}

\begin{table}[t]\renewcommand\arraystretch{1.0}
\begin{threeparttable}[b]
\label{acc}
\caption{Detection rate, model parameters, and number of operations of CB-ConvLSTM, vanilla ConvLSTM, and CNN models}
\begin{center}
\setlength{\tabcolsep}{1.6mm}{
    \begin{tabular}{cc|cccc|c|c}
\hline\hline
&&\multicolumn{4}{c|}{CB-ConvLSTM} &  ConvLSTM  & CNN \\
\multicolumn{1}{c}{$\theta$}        &           & 0.5   & 0.2    & 0.1       & 0& -& -            \\\hline
\multirow{3}{*}{\begin{tabular}[c]{@{}c@{}}Detection \\ rate (\%)\end{tabular}} &$p_{3}$ & 88.50 & 88.50  & 88.70 &   88.88   &88.70 & 57.80\\ 
                                    &$p_{5}$  & 96.70 & 96.76  & 96.90 &   97.07   &97.10 & 77.40 \\
                                    &$p_{10}$ & 99.20 & 99.20  & 99.13 &   99.50   &99.40 & 91.40 \\ \hline \hline
\multicolumn{2}{c|}{ $\#$Parameters (M) } & \multicolumn{4}{c|}{0.42} & 0.42  & 0.40\\ \hline
\multicolumn{2}{c|}{ $\#$Flops (M) }      & 9.00 &9.49 & 10.22 & 18.86 & 42.61 & 18.40\\ 
\hline
\end{tabular}}
\end{center}
\end{threeparttable}
\vspace{-8mm}
\end{table}

\section{Experimental Setup \& Results}

The accuracy of pupil detection is assessed by the Euclidean distance between the predicted pupil center and the ground truth. Distances shorter than $p$ pixels are regarded as successful in the estimation of the detection rate \cite{eivazi2019improving}. In this context, $p_3$, $p_5$, and $p_{10}$ represent $p=3$, $5$, and $10$, respectively.

In the training phase, the Mean Squared Error (MSE) is employed as the loss function, and model parameters are updated using the Stochastic Gradient Descent (SGD) optimizer. A learning rate of 0.001 is maintained over 30 epochs with a batch size of 16. The data allocation for training and validation follows an $80/20$ proportion. We also augment the dataset by using a stride of 1 through event-based streams, thus generating multiple overlapping clips from a single event video stream, each displaced by one frame.

\begin{figure}[t]
    \centering
    \includegraphics[width=0.35\textwidth]{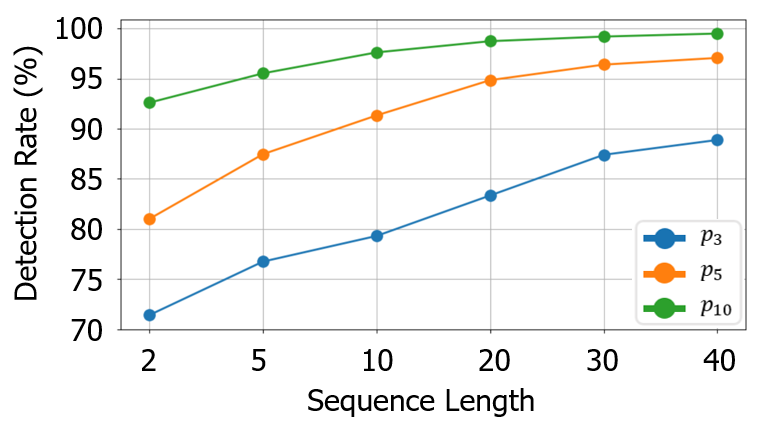}
    \caption{Detection rate under different sequence length}
    \label{fig:seq}
    \vspace{-2mm}
\end{figure}

\begin{figure}[t]
    \centering
    \includegraphics[width=0.35\textwidth]{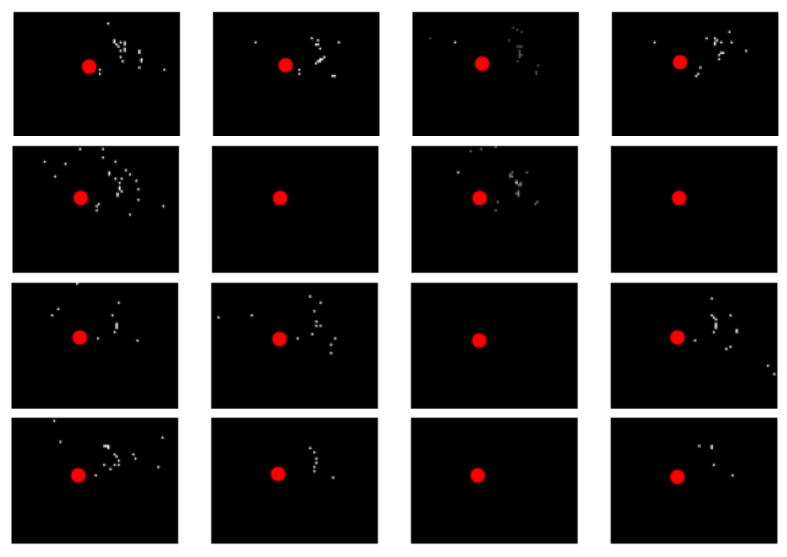}
    \caption{Examples of 16 continuous event-based frames with predicted results. The predicted result is shown as a red dot. Even if few events are generated, our approach can still detect the location of the pupil.}
    \label{fig:label_eventeye}
    \vspace{-4mm}
\end{figure}

Owing to the spatial and temporal sparsity of event streams, many frames often contain minimal information. To address this issue, our eye tracking model integrates a recurrent ConvLSTM structure. As illustrated in Fig.~\ref{fig:seq}, the sequence length in the temporal dimension significantly impacts the detection rate in the ConvLSTM-based model. The detection rates, indicated by $p_3$, $p_5$, and $p_{10}$, raise to 88.8\%, 97.0\%, and 99.5\%, respectively, representing an increase of 17.4\%, 15.8\%, and 6.9\% when the sequence length is increased from 2 to 40. The extension in sequence length aids in acquiring more temporal information, thereby increasing the detection rate, especially when detecting those frames with less information (see Fig.~\ref{fig:label_eventeye}).

The proposed CB-ConvLSTM structure was further evaluated for its contribution towards network sparsity and its impact on the detection rate, and a comparison was made with the vanilla ConvLSTM and CNN models. Table~\ref{sparsity} presents the sparsity levels of various layers and components within the proposed CB-ConvLSTM and the standard ConvLSTM. Table II highlights the detection rate, model parameters, and the number of operations performed by the CB-ConvLSTM at different $\theta$ values, as well as the standard ConvLSTM and a CNN model.

By increasing $\theta$ from $0$ to $0.5$ to explore a broader design space, the temporal sparsity increases from 69.2\% to 85.3\% without adversely affecting the model's precision. Compared with the standard ConvLSTM, the CB-ConvLSTM demonstrates about 3$\times$ greater sparsity. The comparison CNN model, equipped with similar parameters as the ConvLSTM network and composed primarily of six convolution layers with 32, 32, 32, 64, 64, and 128 filters respectively, yields a sub-optimal detection rate ($p_3$ = 57.80\%).

Regarding computational operations, the CB-ConvLSTM performs 8.8X fewer calculations during convolutions, which leads to a 4.7$\times$ reduction in computations for the entire network compared with the standard ConvLSTM and merely half the computational effort demanded by the CNN model.

\section{Conclusion}
This work proposes a computationally efficient event-based eye tracking solution, specifically targeting pupil center detection. By leveraging the ConvLSTM network, we capture the spatio-temporal sparse features inherent in event streams more effectively. Moreover, to reduce computational demands, we introduce the novel CB-ConvLSTM structure to induce high temporal sparsity in activations to reduce arithmetic operations in convolutions. Results indicate that our approach achieves a network sparsity of 85.3\%, facilitating a 4.7$\times$ reduction in arithmetic operations, all while maintaining accuracy, relative to the conventional ConvLSTM-based model. CB-ConvLSTM has the potential to be implemented on specialized hardware exploiting spatio-temporal sparsity~\cite{chen2020efficient, gao2022spartus} to further reduce the inference latency and energy cost.

\newpage
\bibliographystyle{IEEEtran}
\bibliography{refs.bib}

\vspace{12pt}

\end{document}